\documentclass[letterpaper, 10 pt, conference]{ieeeconf}
\IEEEoverridecommandlockouts
\usepackage{amsmath,amssymb,amsfonts}
\usepackage{multirow}
\usepackage{algorithmic}
\usepackage{graphicx}
\usepackage{textcomp}
\usepackage{gensymb}
\usepackage{subcaption}
\usepackage[font=small]{caption}
\usepackage{glossaries}
\usepackage{xcolor}
\def\BibTeX{{\rm B\kern-.05em{\sc i\kern-.025em b}\kern-.08em
    T\kern-.1667em\lower.7ex\hbox{E}\kern-.125emX}}

\usepackage[sorting=none, doi=false,isbn=false,url=false,eprint=false]{biblatex} 
\usepackage{csquotes}

\usepackage{tikz}
\usepackage{textcomp}
\usepackage{hyperref}
\usepackage{lipsum}

\newcommand\copyrighttext{%
  \footnotesize \textcopyright 2024 IEEE. Personal use of this material is permitted.  Permission from IEEE must be obtained for all other uses, in any current or future media, including reprinting/republishing this material for advertising or promotional purposes, creating new collective works, for resale or redistribution to servers or lists, or reuse of any copyrighted component of this work in other works.}
\newcommand\copyrightnotice{%
\begin{tikzpicture}[remember picture,overlay]
\node[anchor=south,yshift=10pt] at (current page.south) {\fbox{\parbox{\dimexpr\textwidth-\fboxsep-\fboxrule\relax}{\copyrighttext}}};
\end{tikzpicture}%
}

\newglossaryentry{CD}{
	name={CD}, 
	description={Chamfer-L1 distance},
	first={Chamfer-L1 distance (CD)}
}
\newglossaryentry{CNN}{
	name={CNN}, 
	description={CNN},
	first={Convolutional Neural Network (CNN)}
}
\newglossaryentry{DOF}{
	name={DOF}, 
	description={DOF},
	first={Degrees of Freedom (DoF)}
}
\newglossaryentry{EMG}{
	name={EMG}, 
	description={electromyographic},
	first={electromyographic (EMG)}
}
\newglossaryentry{VHACD}{
	name={v-HACD}, 
	description={Hierarchical Approximate Convex Decomposition},
	first={Hierarchical Approximate Convex Decomposition (v-HACD)}
}
\newglossaryentry{GSR}{
	name={GSR}, 
	description={GSR},
	first={Grasp Success Rate (GSR)}
}
\newglossaryentry{INSTR}{
	name={INSTR}, 
	description={INSTR},
	first={ Instance Stereo Transformer (INSTR)}
}
\newglossaryentry{IOU}{
	name={IoU}, 
	description={Intersection over Union},
	first={Intersection over Union (IoU)}
}
\newglossaryentry{SGM}{
	name={SGM}, 
	description={SGM},
	first={Semi-Global Matching (SGM)}
}
\newglossaryentry{SCT}{
	name={SCT}, 
	description={Shared Control Template},
	first={\glsentrydesc{SCT} (\glsentrytext{SCT})},
	plural={SCTs},
	firstplural={Shared Control Templates (\glsentryplural{SCT})}
}
\newglossaryentry{SC}{
	name={SC}, 
	description={Shared Control},
	first={Shared Control (SC)},
}
\newglossaryentry{UOIS}{
	name={UOIS}, 
	description={UOIS},
	first={Unknown Object Instance Segmentation (UOIS)}
}
\newglossaryentry{URDF}{
	name={URDF}, 
	description={Unified Robot Description Format},
	first={Unified Robot Description Format (URDF)}
}
\newglossaryentry{VQDIF}{
	name={VQDIF}, 
	description={VQDIF},
	first={Vector Quantized Deep Implicit Function (VQDIF)}
}

\title{\LARGE \bf
Unknown Object Grasping for Assistive Robotics
}

\author{Elle Miller$^\dagger$, Maximilian Durner, Matthias Humt, Gabriel Quere,  Wout Boerdijk, Ashok M. Sundaram, \\ Freek Stulp, and Jörn Vogel
\thanks{All authors were with the DLR Institute of Robotics and Mechatronics during the project. $^\dagger$Elle Miller is now affiliated with the University of Edinburgh. 
This work is partly supported by the Bavarian Ministry of Economic Affairs, Regional Development and Energy (StMWi) by means of the project SMiLE2gether (LABAY102). Contact: {\tt\small elle.miller@ed.ac.uk}}
}

\begin{document}

\makeatletter

\maketitle

\copyrightnotice
\thispagestyle{empty}
\pagestyle{empty}

\begin{abstract} 
We propose a novel pipeline for unknown object grasping in shared robotic autonomy scenarios. State-of-the-art methods for fully autonomous scenarios are typically learning-based approaches optimised for a specific end-effector, that generate grasp poses directly from sensor input. In the domain of assistive robotics, we seek instead to utilise the user's cognitive abilities for enhanced satisfaction, grasping performance, and alignment with their high level task-specific goals. Given a pair of stereo images, we perform unknown object instance segmentation and generate a 3D reconstruction of the object of interest. In shared control, the user then guides the robot end-effector across a virtual hemisphere centered around the object to their desired approach direction. A physics-based grasp planner finds the most stable local grasp on the reconstruction, and finally the user is guided by shared control to this grasp. In experiments on the DLR EDAN platform, we report a grasp success rate of 87\% for 10 unknown objects, and demonstrate the method's capability to grasp objects in structured clutter and from shelves. 
\end{abstract}

\section{Introduction}
\label{sec:introduction}
The ability to grasp a wide variety of objects is crucial for independent living. 
Daily activities such as eating, drinking, dressing, grooming, and others all rely on the underlying capability to retrieve and interact with objects. 
This ability is often compromised in individuals with motor impairment, which can arise from spinal cord injury, traumatic brain injury, multiple sclerosis, muscular atrophy, and various other neuromuscular diseases. 
A review of surveys, reflecting the views of over 200 people with motor impairments, found that tasks relating to picking up, fetching, and carrying objects were of highest priority for an assistive technology \cite{priority}. 
There is a large body of research concerning the grasping of \textit{known} objects, for which full object information and models are available to assist with grasp planning. 
For assistive technologies however, the first application environment is the home, which may have thousands of items.
In a given day, an individual may need to interact with food, clothes, hygiene products, electronics, personal belongings, appliances, kitchenware, all of which are likely to vary and update through time. 
The ability to grasp only known objects greatly reduces the potential applications, and thus the focus of this work is the grasping of completely unknown rigid objects.

This problem has seen recent success with deep neural networks that can generate grasp poses directly from RGB or point cloud input, e.g. \cite{sundermeyer_contact-graspnet_2021}.
While fully autonomous methods could be applied in this setting, we argue that user satisfaction would be lower. A study with 10 traumatic spinal cord injured subjects \cite{autonomy} found that autonomy did not provide more satisfaction for a robotic pick-and-place task. The authors conclude that there is a need to appropriately channel the autonomy provided, to enhance user satisfaction: 

\begin{displayquote}

\textit{``While able-bodied users may prefer to cede autonomy to robots, we believe that disabled users tend to see the robot not merely as an agent to retrieve objects but also as a quintessential tool to reassert their domain of interaction with their environment as well as engage and exercise their cognitive faculties to the fullest."}
\end{displayquote}

\begin{figure}[t!]
    \centering
    \includegraphics[width=0.9\linewidth]{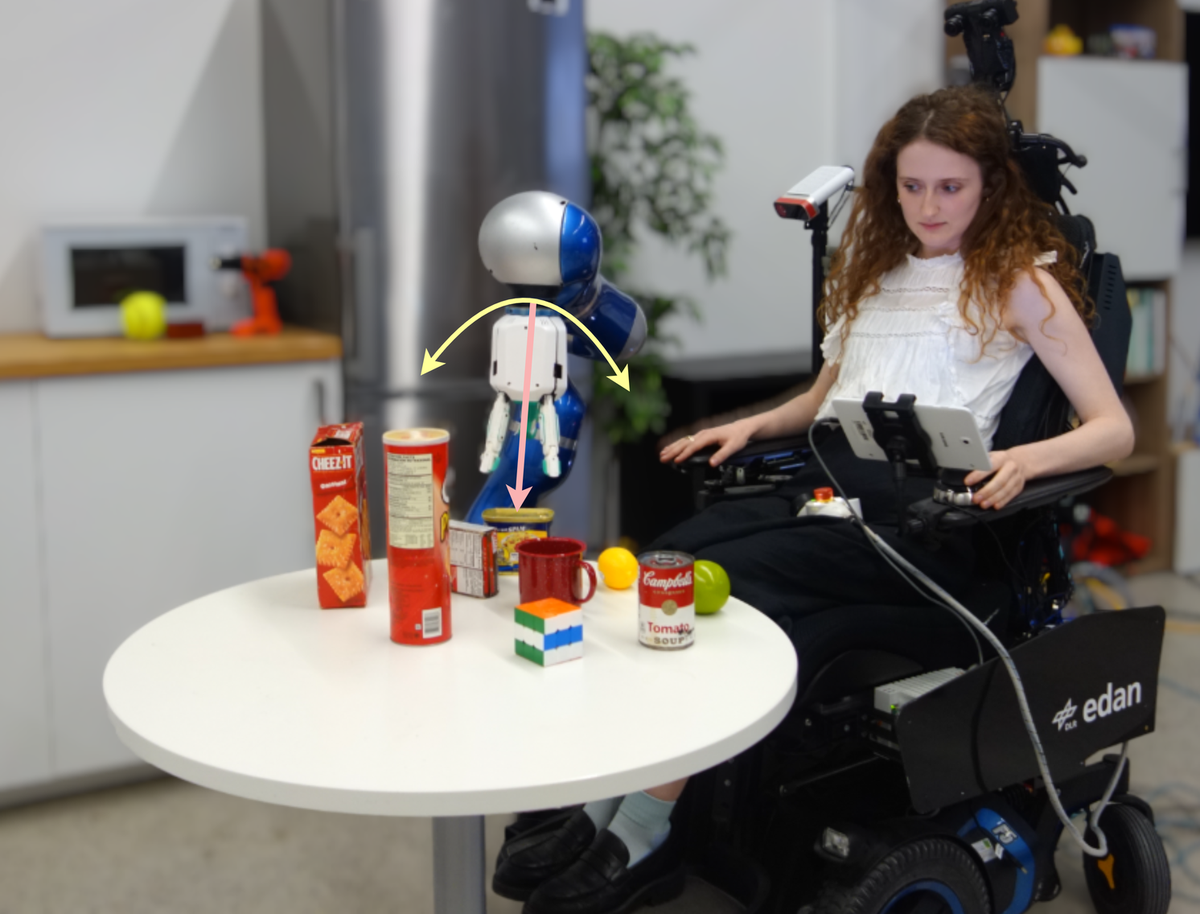}
    \caption{Grasping an unknown object in structured clutter with the DLR EDAN assistive robotic system. After selecting an object, the user guides the end-effector across a virtual hemisphere to their desired approach direction. Depicted is an author, not a target user.\vspace{-0cm}}
    \label{fig:sc}
\end{figure}

Moreover, the user already has intuition of where the best grasp might be, and they may wish to align the grasp with high level task-specific goals, e.g. transporting an empty mug compared to drinking from a full mug. In light of these findings, we aim to design a system that (i) requires the user to actively interact with the environment and (ii) utilises their cognitive abilities for decision-making. Yet, using manual control to perform goal-directed multi-fingered object manipulation is complex and difficult due to the high number of \gls{DOF} in robotic manipulators. 
Hence, our system uses shared control so the user can more intuitively control the end-effector \cite{sct}\cite{siciliano2008springer}.

\begin{figure*}[t!]
    \centering
    \includegraphics[width=0.8\textwidth]{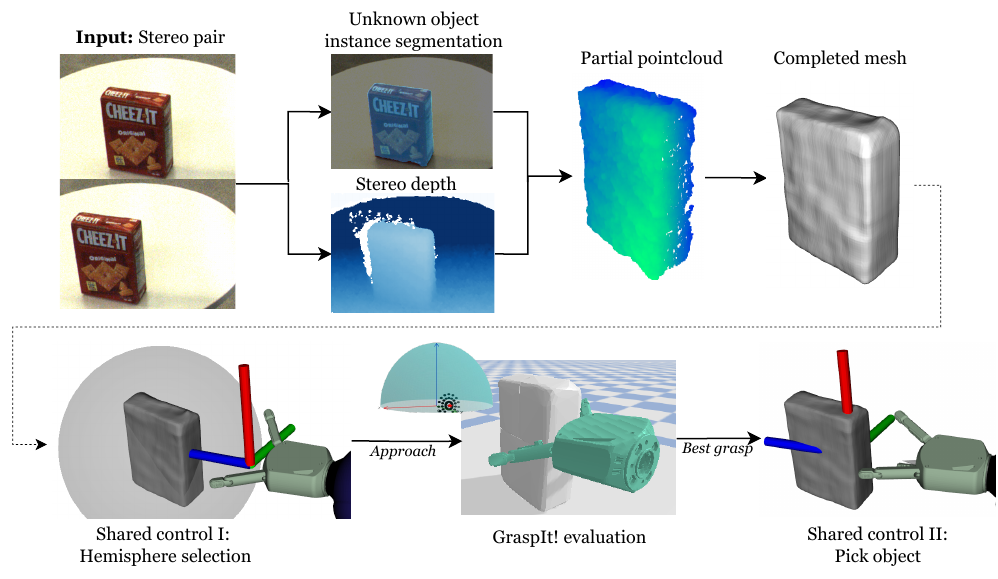}
    \caption{Our unknown object grasping pipeline. The perception module transforms a pair of rectified stereo images to the completed mesh of an unknown object. Unknown object instance segmentation is performed using INSTR \cite{instr}, and for shape completion we use the method presented in~\cite{humt2023_shape} which applies a \gls{VQDIF}~\cite{yan2022shapeformer} trained on simulated Kinect data. In shared control \cite{sct}, the user guides the end-effector across a virtual hemisphere centered around the unknown object, selects their desired approach position, and a physics-based grasp planning simulation finds the best local grasp at which the actual grasp is executed automatically.\vspace{-0.45cm}}
    \label{fig:system}
\end{figure*}

Our pipeline (Fig.~\ref{fig:system}) performs \gls{UOIS} on a pair of stereo images, and generates a 3D mesh reconstruction of the object using a deep implicit function network. 
The user then guides the robot end-effector across a virtual hemisphere centered around the unknown object, and selects their desired approach position. 
Next, a physics-based grasp planning simulation is run on the reconstructed mesh to find the most stable grasp near the user's selected approach.  Finally, the user is guided by shared control to the planner's best grasp pose.

The main advantages of the pipeline are outlined below.
\begin{enumerate}
  \item[i)] \textit{Human-centred:} The user is established as an active cognitive agent for enhanced satisfaction, grasping performance, and alignment with task-specific goals.
  \item[ii)] \textit{End-effector agnostic:} There is no additional time or effort cost  for changing or altering the end-effector. 
  \item[iii)] \textit{Customisable:} The number and region of local grasps sampled can  be adjusted according to user ability.
  \item[iv)] \textit{Scalable:} The system can be extended to more complex scenes, relying on human intuition to overcome difficulties (e.g. collision-avoidance, Fig.~\ref{fig:sc}\&\ref{fig:shelf}). 
\end{enumerate}

The novel contribution of this work is a pipeline for grasping unknown objects in shared autonomy or assistive settings, that utilises the valuable cognitive abilities of the user. We provide experimental verification of the method on the DLR EDAN assistive robotic system with the CLASH hand \cite{edan}\cite{clash}, showing high grasping success rates for singulated rigid objects and extensions to more complex scenes.

\section{Related Work}
\label{sec:background}
\subsection{Robotic grasping of unknown objects}
\label{sec:related_grasping}
Robotic grasping is the task of planning and executing grasp poses and associated finger configurations for a given object and end-effector. Analytic methods directly analyze the object geometry, and data-driven methods typically employ machine learning. Many learning-based methods predict planar grasps (2D
position, 1D rotation), suited for scenarios like bin picking \cite{pinto2015supersizing} \cite{Mahler2017LearningDP}. In complex environments however, such as the home,
additional task and collision constraints make 6D grasp prediction crucial.  
\cite{newbury_deep_2023} review the application of deep learning to 6D grasp pose synthesis, identifying the major methods to be sampling, regression, and reinforcement learning (RL). Sampling methods learn a function to evaluate the quality of a sampled grasp \cite{varley}, regression (or end-to-end) methods learn a function to predict high-quality grasps from visual information \cite{sundermeyer_contact-graspnet_2021}\cite{fang_anygrasp_2023}, and RL aims to learn a policy that maximises total reward \cite{levine2016learning}\cite{pmlrkalashnikov2018qtopt}.

We did not explicitly use one of these approaches, in order to meet two criteria: the incorporation of \textit{user intent} and scalability to \textit{complex end-effectors}. As end-to-end methods simultaneously reason on perception, grasp planning, and control, it was unclear how to integrate a user's input for further reasoning, which we deemed to be essential for this shared-autonomy system as discussed in Section~\ref{sec:introduction}. Since these systems can not provide feedback on individual modules, this may also potentially result in lower user satisfaction and acceptance. We considered extending approaches such as Contact-GraspNet \cite{sundermeyer_contact-graspnet_2021} to generate grasp candidates that the user could choose between, but found that many works were heavily optimised for simple rigid end-effectors. For example, \cite{sundermeyer_contact-graspnet_2021} exploit the symmetry of a parallel-jaw gripper to formulate a simple loss function and reduce the problem dimensionality from 6D to 4D. These methods can not be trivially replicated for complex end-effectors such as our CLASH \cite{clash}, an intrinsically compliant 7 \gls{DOF} hand inspired by the tendon routing in humans. Moreover, end-effectors may be updated through time for different requirements (e.g. EDAN previously used the DLR-HIT hand \cite{liu_multisensory_2008}). Thus, learning-based methods typically require problem reformulation, revised data collection, and retraining for different end-effectors. 
While it may appear that our proposed pipeline is complex with several failure points, we note that many other approaches also rely on RGB and pointcloud input, e.g. \cite{sundermeyer_contact-graspnet_2021}\cite{fang_anygrasp_2023}. Finally, the recent realisation of semantic manipulation e.g. \textit{``Grab the left ear of the elephant"} \cite{sundaresan_kite_2023} provides interesting opportunities for human-centred grasping. However, we believe our method offers finer-grained grasp selection and increased satisfaction through active interaction for users in assistive settings.

\subsection{Shape completion enabled grasping}
Reconstructing a 3D model of an object using only a partial point cloud is a promising method for enabling robotic grasping. \cite{varley} use a 3D \gls{CNN} for shape completion, trained on simulated depth images from 484 objects and achieve a 93\% grasp success rate (GSR) on 15 YCB objects, most of which were already used during training. \cite{chen} propose a transformer-based encoder for shape completion, trained on real depth images from the YCB-Video dataset and achieve an 83\% GSR across 6 YCB objects, all of which were seen during training. 
\cite{3ds} propose a transformer-based encoder-decoder network for shape completion trained on the data from \cite{varley}, and achieve a 76\% GSR across 10 YCB objects, which also have been used at least partially during training. 
\cite{humt2023_shape} recognise the challenges of reconstruction from partial and noisy depth data, and train an implicit function network \cite{yan2022shapeformer} on simulated Kinect camera depth images to further bridge the simulation gap to real sensing conditions. 
We implement this method of shape completion in our work, and refer the reader to \cite{humt2023_shape} for further information and related work. 
In order to recover an unknown object point cloud from a scene, we initially perform \gls{UOIS}. 
In this context, \cite{instr} present a novel stereo-based transformer approach, addressing corrupted depth maps in real-world scenarios. 
To our knowledge, this is the first work combining RGB-based \gls{UOIS} with shape completion for robotic grasping.

\subsection{Assistive robotic grasping}

Shared control grasping in assistive robotic systems has been studied previously. 
\cite{muelling_autonomy_2017} focus on known objects with predefined grasp poses and approach vectors.
\cite{vogel2016flexible} use a dense grasp database to provide the user with more freedom in selection, but also for known objects. Most similar to our work, 
\cite{jain2016grasp} study grasping for unknown objects, but their approach is limited to objects that fall into shape primitive categories. Their system autonomously selects a single grasp for execution from semantic candidates (e.g. top, side, pinch) based on the user's teleoperation of the end-effector. 
While this method also incorporates user intent, we believe our approach allows greater unknown object generalisation and user autonomy through finer-grained grasp direction selection beyond pre-defined semantic categories.

\section{Method}
\label{sec:method}
Our proposed pipeline covers all relevant modules -- from detecting object instances up to the prediction of robust grasp poses -- for unknown object grasping. 
Given a pair of stereo images, the goal of perception is to create a complete mesh for each object in the scene (see upper part of Fig.~\ref{fig:system}).
One should note that the use of mesh reconstruction limits this method to rigid objects.
To this end, we perform \gls{UOIS} and employ \gls{INSTR}~\cite{instr} that creates pixel-wise instance masks for all arbitrary objects visible in the current view.
Next, for higher robustness, each of the binary segmentation masks is filtered by contour area for noise. 
The user is presented with the filtered instances, and they select a target object by cycling through the options. 
We then construct an instance-specific point cloud by masking the \gls{SGM}~\cite{hirschmuller2005accurate} created depth map with the binary mask of the object of interest.
After down sampling and outlier removal, the point cloud is reconstructed using the method presented in~\cite{humt2023_shape} which applies a \gls{VQDIF}~\cite{yan2022shapeformer} trained on simulated Kinect data.
Finally, we decimate the completed mesh to $\sim$1000 triangular faces using the Quadric Error Metrics \cite{dec} for improved collision computation in the physics simulator.
The completed mesh is then forwarded to the grasping module.
As shown in Fig.~\ref{fig:system}, the grasping module consists of three parts: hemisphere selection, grasp evaluation, and picking up the object. 
For both hemisphere selection and object picking, we employ \glspl{SCT}, previously developed for the EDAN system~\cite{sct}.
An \gls{SCT} maps low-dimensional user input to end-effector motions and provides task space constraints, acting as as regional constraints as defined by \cite{bowyer2013active}, to guide the robots end-effector within the task.
In this case a 3 \gls{DOF} joystick is used to control the end-effector.
The end-effector is constrained on a virtual hemisphere centered around the object of interest, as well as constrained to point towards the hemisphere origin.
Those constraints allow the user to select their desired approach pose:
they can toggle between moving the end-effector across the hemisphere surface (translational control), and rotating the end-effector around its axis pointing towards the hemisphere origin (rotational control). 
When satisfied with the approach pose, the user triggers a button. 

\begin{figure}[t!]
\centering
\begin{subfigure}[t]{0.49\columnwidth}
   \includegraphics[height=2.5cm]{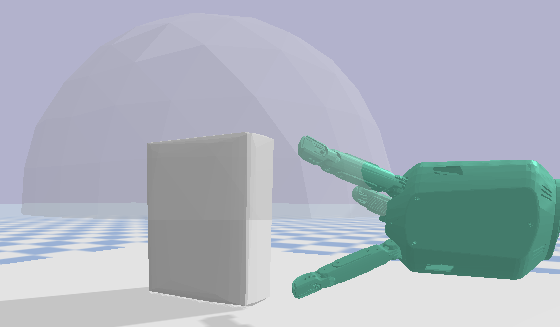}
   \caption{Power grasp hemisphere}
   \label{fig:Ng1} 
\end{subfigure}
\begin{subfigure}[t]{0.49\columnwidth}
   \includegraphics[height=2.5cm]{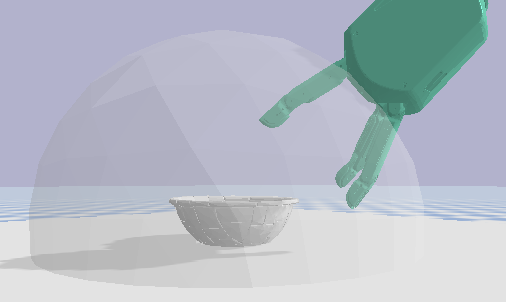}
   \caption{Precision grasp hemisphere}
   \label{fig:Ng2}
\end{subfigure}
    \caption{Grasp approach hemispheres, decided by object height. The power grasp allows for side-on grasps, and the precision grasp allows for greater sampling around the center point.}
    \label{fig:grasp}
\end{figure}

Given this approach pose, we use the grasp planning simulation  GraspIt!~\cite{miller2004graspit} on the reconstructed mesh to find the best local grasp.
For a more realistic physics simulation, we used the PyBullet implementation~\cite{pybullet_graspit} of GraspIt! to exploit the improved Bullet dynamics, kinematics, and collisions. 
We place the object on a surface and apply gravity instead of being fixed in space, to simulate object-surface dynamics. The unknown object is defined by a \gls{URDF} file. 
By default, PyBullet uses a convex hull for mesh collision detection, which means that objects with concavities are (incorrectly) approximated as convex. 
To deal with this, we used the in-built \gls{VHACD}~\cite{vhacd} to decompose the reconstructed mesh into multiple convex hull objects. 
The default mass is set to 0.3 kg. 
The lateral, rolling, and spinning friction coefficients are set to 0.3, 0.01, and 0.01 respectively. 
The inertia is recomputed by Bullet based on mass and volume of the collision shape.
For grasp planning, we define a hemisphere from which approach positions are sampled from. 
We split the grasping strategy into \textit{power} and \textit{precision} grasps, depending on the object height (Fig.~\ref{fig:grasp}).
The power grasp hemisphere is raised by 12 cm to allow for ``side-on" grasps (without our end-effector colliding with the table). 
The original GraspIt! generates grasp approach angles evenly spaced across a sphere but we only sample local positions, which drastically reduces the amount of sampling positions and thus required sampling time. We generate a set of circumferences at different angular offsets from the user's pose, and sample points from them. 
An example sampling is illustrated in Fig.~\ref{fig:system} on the blue hemisphere in grasp evaluation, with 3 circumferences evenly spaced between a 0-10$\degree$ angular offset from the user's approach pose, which is shown in red. 
Note that the approach poses which exceed the hemisphere surface are filtered out. 
The final parameter we sample is how bent the fingers are on approach, which we call \textit{finger flexion}: a value of 0 is where each finger is fully outstretched, and a value 1 corresponds to completely closed fingers.
For power grasps, we sample finger flexions \{0, 0.1\} per approach pose. 
For precision grasps, finger flexion is much more consequential to grasp success, and as such we sample \{0.1, 0.2, 0.25, 0.3, 0.35\}. 

\begin{figure}[t!]
    \centering
    \includegraphics[width=0.9\linewidth]{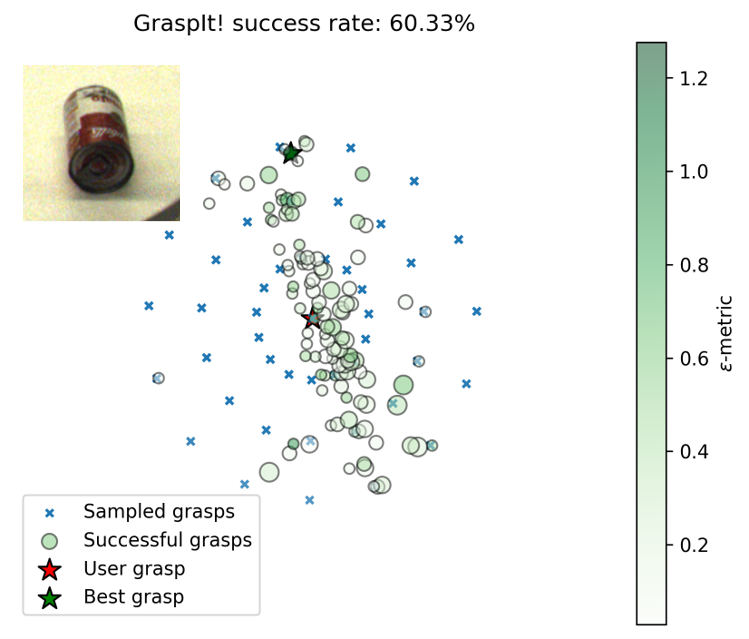}
    \caption{Visualisation of GraspIt! simulation output for a soup can. From the user's selected approach (top-down), local approach poses are sampled from 3 circumferences spaced between 0-10$\degree$ from the input (blue crosses) projected onto the hemisphere. The hand approaches the soup from these poses but will keep moving until 3 contacts are reached, causing it to envelope around the object. Successful grasps are shown in circles, where the circle size is proportional to finger flexion (i.e. the smallest circles correspond to finger flexion 0.35, largest 0.1), and colour shows grasp quality.}
    \label{fig:soup}
\end{figure}

For each approach pose and finger flexion, the hand moves towards the center of the hemisphere until it has sufficient contact points (2 with the object for power grasps, 3 with object and/or world for precision grasps). 
Note that the object is fixed here. The position of the hand with a small ``back-off" margin applied is then saved. The hand is then fixed, and the object is unfixed. Gravity is applied to the object, and the hand is closed with velocity control. 
Finally, the surface is moved vertically downwards. If there are hand-object contact points after 3 seconds, the grasp pose is deemed successful. 
Finally, the grasp with the highest $\epsilon$ value is selected (Fig.~\ref{fig:soup}.
If no grasp is found, the user is asked to select another approach position. 
The  $\epsilon$ quality measure is the radius of the largest sphere that fits in the convex hull of the Grasp Wrench Space, and represents how robust the grasp is to external disturbance \cite{miller2004graspit}. 
We calculate $\epsilon$ of the grasp pose at 0.5 seconds after the surface is moved rather than at 3 seconds, as a more accurate representation of grasp quality. 

\section{Results}
\label{sec:results}
We implement our method on EDAN, the EMG-controlled Daily AssistaNt, a robotic wheelchair mounted with a torque-controlled, 8 \gls{DOF} version of the DLR lightweight robot (LWR-III) \cite{edan}. 
The robot can be controlled by a joystick or via \gls{EMG} signals, and is designed to perform activities of daily living. 
We evaluate our method using 3 DOF joystick control, operated by one of the authors. Prior research shows that people with severe motor impairment can perform three-dimensional tasks from neural signals \cite{hochberg_reach_2012} or EMG \cite{hagengruber_functional_2018}. Thus, we hypothesise that our method will also work with target users requiring alternative interfaces, but with longer control times. We believe that accuracy can be maintained if the sampling scheme is customised according to the precision abilities of the user, e.g. sample 0-30$\degree$ from the selected approach for users with less control.

For stereo vision we use 2x Point Grey Chameleon CMLN-13S2C-CS cameras with 1/1.8" 4.5mm C-Mount wide angle lenses. 
For grasping we use the CLASH hand, a 7 \gls{DOF} Compliant Low-cost Antagonistic Servo Hand also developed at DLR~\cite{clash}. 
Our GraspIt! implementation is parallelised and run on a computer with 55 cores. 
For grasping, we select 10 objects  from the YCB dataset~\cite{ycb}.
The primary experiment is grasping singulated objects on a surface.
Additionally, we compare autonomy modes and demonstrate the capability to grasp objects in structured clutter and from a shelf.
We do not compare our results against baselines, e.g. Contact-GraspNet \cite{sundermeyer_contact-graspnet_2021}, because most related works have been trained on or optimised for parallel-jaw grippers, which would be non-trivial to re-implement for the CLASH hand (see Section~\ref{sec:related_grasping}).
\subsection{Singulated objects}
For each of the 10 YCB objects we run 10 grasping trials, where the object's position and orientation are randomised within an area the size of an A4 sheet of paper. 
A trial is successful if the object remains in the air 5 seconds after being grasped, lifted, and held stationary. 
The experimental results are presented in Table~\ref{tab:edan}. 
To evaluate \gls{INSTR}, the mean \gls{IOU} and standard deviation for each object is provided. 
The shape completion metrics we use are the volumetric \gls{IOU} and \gls{CD} mean and standard deviation. 
The mean GraspIt! success rate refers to the percentage of approaches that maintained object contact 3 seconds post gravity in simulation.
Finally, the \gls{GSR} is the fraction of successful physical trials. 
To test the effectiveness of the complete pipeline, we only proceeded with attempts where a reconstructed mesh was recovered. 
As such, we discarded 5 attempts (2 \textit{Banana}, 3 \textit{Strawberry}) where \gls{INSTR} failed to detect or severely under-segmented the object.

\subsection{Autonomy modes}
We show the benefits of our pipeline in an assistive setting and compare our approach (SC+GraspIt!) against varying degrees of autonomy (Table~\ref{tab:manual}). 
In manual mode, the user can toggle between controlling the end-effector translation, rotation, and closing the gripper. 
In \gls{SC} only mode, we remove GraspIt! from the pipeline and the user instead guides the end-effector along a line segment between their approach pose pose and the hemisphere origin, and manually triggers the gripper closing.

\setlength{\tabcolsep}{3.5pt}
\begin{table}[h!]
\caption{Comparison of results with different levels of autonomy.}
\begin{center}
\begin{tabular}{ccccccccc}
\hline
\multirow{2}{*}{\textbf{Autonomy}} & \multicolumn{2}{c}{\textbf{Cracker Box}} & \multicolumn{2}{c}{\textbf{Strawberry}} & \multicolumn{2}{c}{\textbf{Bowl}} & \multicolumn{2}{c}{\textbf{Mean}}\\
& Time & GSR & Time & GSR & Time & GSR &  Time & GSR\\
\hline
\hline
Manual & 62 & 90\% & 50 & 90\% & 57 &  100\% & 56 & 93\% \\
SC & 36 & 100\% & 33 & 60\% & 43 & 30\% & 37 & 63\%\\
SC+GraspIt! & 41 & 90\%  & 36 & 80\%  & 55 & 60\% & 44 & 77\% \\
\hline
\end{tabular}
\vspace{-0.3cm}
\label{tab:manual}
\end{center}
\end{table}

\begin{figure}[t!]
    \centering
    \includegraphics[width=\linewidth]{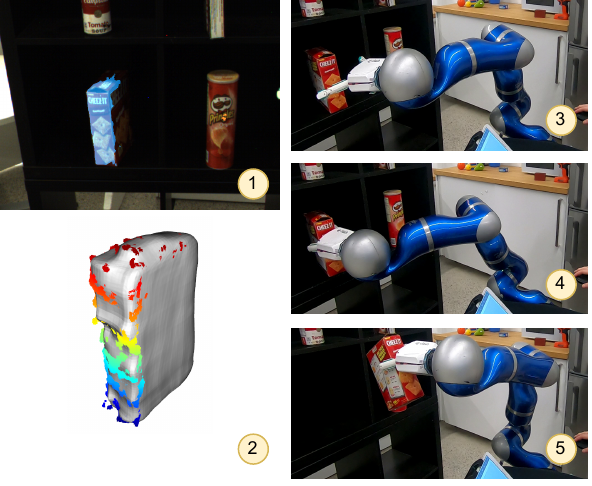}
    \caption{Shelf grasping experiment. 1. INSTR prediction 2. Completion prediction 3. User's selected hemisphere approach 4. GraspIt! predicted grasp 5. Successful grasp. More complex grasping scenarios utilise the user's intuition for collision avoidance.\vspace{-0.1cm}}
    \label{fig:shelf}
    \vspace{-0.3cm}
\end{figure}

\subsection{Complex scenes}
We also demonstrate the scalability of our pipeline to more complex scenes.
In structured clutter (Fig.~\ref{fig:sc}), we grasped 5/5 objects (\textit{Spam tin}, \textit{Lemon}, \textit{Pringles}, \textit{Pear} and \textit{Cracker box}) from the scene on the first trial. 
From the shelf (Fig.~\ref{fig:shelf}), we demonstrate grasping of the \textit{Cracker box} despite degraded visual conditions. Both these scenes relied on the user's intuition to select collision-free grasp directions for success.

\section{Discussion}
\label{sec:discussion}

\subsection{Singulated objects}
\begin{figure}[t!]
    \centering
    \includegraphics[width=0.95\linewidth]{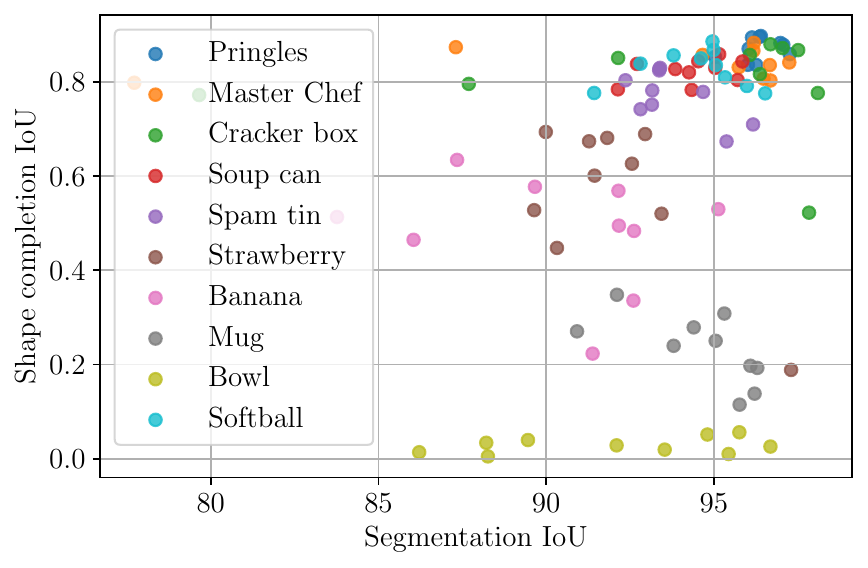}
    \caption{Segmentation and completion IoU for the 100 trials of singulated objects.}
    \label{fig:scatter}
    \vspace{-0.1cm}
\end{figure}

From Table~\ref{tab:edan}, we see that the \gls{UOIS} method performed well with all objects having a mean \gls{IOU} above 90. 
The applied shape completion approach is able to faithfully reconstruct most objects, but struggles with small ones (\textit{Strawberry}, \textit{Banana}), as they are relatively more affected by imprecise segmentation and noise. The volumetric \gls{IOU} metric is deceptively small for objects with small volume (\textit{Mug}, \textit{Bowl}), as even small pose discrepancies can lead to negligible overlap. 
The \gls{CD} metric is not affected by this phenomenon and provides a more faithful representation of shape completion performance in these cases. 
Comparing \textit{Spam tin} with an \gls{IOU} of 77.2 and CD of 0.325 to \textit{Bowl} with an  \gls{IOU} of 2.8 and CD of 0.379, we observe a significant difference in  \gls{IOU} but only a modest increase in CD. 
To see how segmentation performance impacts the shape completion, we look to Fig.~\ref{fig:scatter}. 
There appears to be no clear correlation between segmentation \gls{IOU} and shape completion  \gls{IOU}, except for a dense cluster of high \glspl{IOU} in the upper right. 
This suggests that the shape completion model is robust to minor under- or over-segmentation. 
The GraspIt! success rate is moderately stable across objects, except the \textit{Strawberry}, \textit{Banana}, and \textit{Bowl}. This indicates that for these objects the finger flexion parameter is very influential to success, where only small finger spreads during approach led to successful grasps.
Considering run time, we can see that the major time is taken by the user to select a satisfactory approach pose (24s). 
Note that the timings do not include perception and object selection, which took $<$10s.  

\begin{table*}[t!]
\caption{Experimental results on robot for 100 grasping trials of singulated objects. For each object, the mean and variance across 10 trials are presented for the segmentation IoU, completion volumetric IoU and Chamfer-L1 distance. The mean GraspIt! simulation success rate, and grasp pipeline times are also reported. The GSR is the physical trial result.}
\begin{center}
\begin{tabular}{ccccccccccccc}
\hline
\multirow{3}{*}{\textbf{YCB Object}} & \multicolumn{2}{c}{\textbf{Segmentation}} 
 & \multicolumn{4}{c}{\textbf{Completion}} & \multicolumn{1}{c}{\textbf{GraspIt!}} &  \multicolumn{3}{c}{\textbf{Timings}} & \multirow{3}{*}{\textbf{GSR (10 trials)}} \\
 & \multicolumn{2}{c}{\textbf{IoU}} & \multicolumn{2}{c}{\textbf{IoU}} & \multicolumn{2}{c}{\textbf{CD$\times$10}} &  \multirow{2}{*}{Success rate (\%)} & 
 \multirow{2}{*}{Select} & \multirow{2}{*}{GraspIt!} & \multirow{2}{*}{Pick} & \\
 & $\mathbf{\bar{x}}$ & $\mathbf{\sigma}$ & $\mathbf{\bar{x}}$ & $\mathbf{\sigma}$  & $\mathbf{\bar{x}}$ & $\mathbf{\sigma}$ & &&&&\\
\hline
\hline
Pringles & 96.4 & 0.6 & 87.0 & 2.3 & 0.091 & 0.012 & 59 & 15s & 4s & 10s & 100\% \\
Master Chef can & 93.5 & 6.2 & 83.9 & 3.0 & 0.232 & 0.045 & 72  &  28s & 5s & 12s & 80\% \\
Cracker box & 93.9 & 6.0 & 80.1 & 10.6 & 0.195 & 0.076 & 68  &  29s & 4s & 8s & 90\% \\
Tomato soup can & 94.4 & 1.2 & 82.3 & 2.6 & 0.226 & 0.036 & 70  & 17s & 9s & 11s & 90\% \\
Spam tin & 93.8 & 1.3 & 77.2 & 5.3 & 0.325 & 0.072 & 69   & 23s & 9s & 12s & 100\% \\
Strawberry & 92.1 & 2.2 & 56.5 & 15.7 & 0.672 & 0.188 &  10  &  29s & 6s & 11s & 80\% \\
Banana & 90.3 & 3.5 & 48.2 & 12.1 & 0.300 & 0.103 & 10  &  27s & 6s & 13s & 80\% \\
Mug & 94.6 & 1.8 & 23.4 & 7.3 & 0.527 & 0.219 & 57  &  22s & 9s & 10s & 100\% \\
Bowl & 92.0 & 3.7 & 2.8 & 1.7 & 0.379 & 0.144& 33  & 28s &  7s & 20s & 60\% \\
Softball & 94.5 & 1.5 & 82.8 & 3.9 & 0.314 & 0.067 & 57  &  22s & 7s & 10s & 90\% \\
\hline 
\textbf{Averages} & && & & & & & \textbf{24s} & \textbf{7s} & \textbf{12s} & \textbf{87\%} \\
\hline
\end{tabular}
\vspace{-0.7cm}
\label{tab:edan}
\end{center}
\end{table*}

\begin{figure}[t!]
    \centering
    \includegraphics[width=0.95\linewidth]{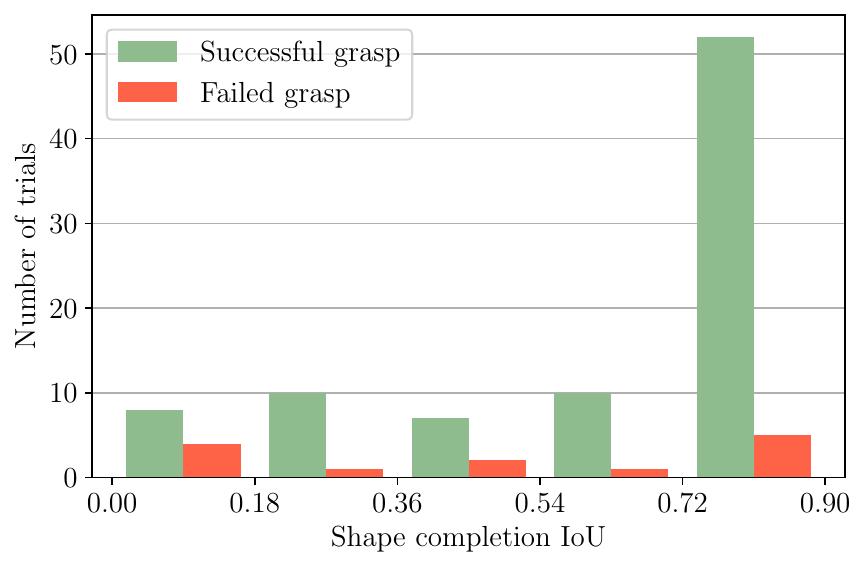}
    \caption{Histogram of shape completion IoU showing number of successful and failed trials for each region.}
    \label{fig:hist}
    \vspace{-0.1cm}
\end{figure}

In the physical trials, we see reliable grasping across many objects. 
The worst performing object was the \textit{Bowl} with a 6/10 GSR, followed by the \textit{Banana}, \textit{Strawberry}, and \textit{Master Chef can} with 8/10 GSR. 
Therefore, does poor shape completion lead to grasp failure? 
Not necessarily, as in Fig.~\ref{fig:hist} we see that grasp failures and successes occurred throughout entire completion \gls{IOU} range.
Most successful grasps had a high shape completion \gls{IOU} ($>$80), but there were also many successful grasps at low completion IoUs. 
For example, the \textit{Mug} had a low mean completion \gls{IOU} of 23.4 (likely due to its shiny surface and concavity), but since the overall object perimeter was preserved had 10/10 \gls{GSR} for top-down grasping. 
For the bowl however, an accurate completion was likely very important for grasp planning as a specific approach position and contacts are required for success. 
Thus, object complexity is likely an important confounding factor to both completion IoU and grasping success.

\subsection{Autonomy modes}
The autonomy experiments in Table~\ref{tab:manual} highlight interesting comparisons between time taken and reliability. 
Manual mode was the most reliable grasping method (28/30 GSR), but also took the most time (56s). 
The \gls{SC} only mode was the fastest (37s), but had the lowest reliability (19/30 GSR). 
This is because when approaching the object with outstretched fingers along the hemisphere radius, it was difficult to visualise how the fingers would establish contact with the object at the resulting grasp pose. 
For example, there was a small window in which the strawberry could be grasped: too low and the outstretched fingers would collide with the surface, too high the strawberry would slip. 
Finally, SC+GraspIt! balanced the time and reliability of the other methods. 
Note that we do not include the timing of the perception pipeline ($<$10s) to contrast timings where the user is under assumed cognitive load. 
Considering the full  pipeline, SC+GraspIt! becomes 44+10=54s, which is very close to the manual average of 56s. 
This result, in combination with the higher manual grasping reliability, may appear to undermine our pipeline. 
However, these measurements do not consider the duration spent under cognitive load. The user is assumed to be under high cognitive load throughout the full length of manual mode (56s), but only during the hemisphere select phase in our method (24s), $\sim44\%$ of the full pipeline time. 
Future work should explore methods of improving the grasping performance in order to reach the high reliability of manual mode, and quantify the differences in cognitive load between the approaches, e.g. using the NASA Task Load Index.

\section{Conclusion}
In this work we present a novel pipeline for grasping unknown objects in an assistive setting, combining \gls{UOIS} with shape completion.
Importantly, our grasping methodology utilises the cognitive abilities of the user for enhanced satisfaction, grasping performance, and alignment with high level task-specific goals. 
We report a grasping success rate of 87\% across 10 singulated objects, and demonstrate capability to grasp objects in structured clutter and from shelves. 
We find that the shape completion model is robust to minor segmentation failures, but can struggle with poor sensor data. 
In experiments comparing autonomy modes, we find that our pipeline reduces the time spent in assumed cognitive load by more than half, compared to manual mode.
Future work could explore entire scene reconstruction and integrating collision-avoidance for dense clutter scenarios, and quantifying cognitive load across autonomy modes.

\renewcommand*{\bibfont}{\normalfont\small}
\printbibliography 

\end{document}